\pdfoutput=1  
\documentclass[sigconf, nonacm]{acmart}

\usepackage{multirow}

\usepackage{algorithm}
\usepackage{algorithmic}
\usepackage{subcaption}
\usepackage{placeins}
\usepackage{soul}

\AtBeginDocument{%
  }

\setcopyright{cc}
\setcctype{by}

\makeatletter
\let\ACM@orig@copyrightpermission\@copyrightpermission
\def\@copyrightpermission{%
  Accepted at CIKM~'26.\par\smallskip
  \ACM@orig@copyrightpermission}
\makeatother

\newcommand{\spara}[1]{\smallskip\noindent{\bf #1}}
\newcommand{\sparas}[1]{\vspace{2pt}\noindent{\bf #1}}

\newcommand{\stdevsize}[1]{\scriptsize{#1}}
\newcommand{\valpm}[2]{#1\stdevsize{$\pm$#2}}
\newcommand{\valpmb}[2]{\textbf{#1\stdevsize{$\pm$#2}}}
\newcommand{\valpmu}[2]{\underline{#1\stdevsize{$\pm$#2}}}

\newenvironment{squishlist}
{\begin{list}{$\bullet$}
 {\setlength{\itemsep}{0pt}
     \setlength{\parsep}{3pt}
     \setlength{\topsep}{3pt}
     \setlength{\partopsep}{0pt}
     \setlength{\leftmargin}{1em}
     \setlength{\labelwidth}{1em}
     \setlength{\labelsep}{0.5em} } }
{\end{list}}

\begin{document}

\title{Inductive Correlation Clustering with Graph Neural Networks}

\settopmatter{authorsperrow=2}

\author{Francesco Paolo Nerini}
\orcid{0009-0000-2936-1297}
\affiliation{%
  \institution{Sapienza University of Rome, Italy}
  \country{}
}
\email{nerini@diag.uniroma1.it}

\author{Francesco Bonchi}
\orcid{0000-0001-9464-8315}
\affiliation{%
  \institution{Intesa Sanpaolo AI Research, Italy}
  \country{}
}
\email{francesco.bonchi@intesasanpaolo.com}

\author{Arijit Khan}
\orcid{0000-0002-7312-6312}
\affiliation{%
  \institution{Bowling Green State University, OH, USA}
  \country{}
}
\email{arijitk@bgsu.edu}

\author{André Panisson}
\orcid{0000-0002-3336-0374}
\affiliation{
 \institution{Intesa Sanpaolo AI Research, Italy}
 \country{}
}
\email{andre.panisson@intesasanpaolo.com}

\renewcommand{\shortauthors}{Francesco Paolo Nerini, Francesco Bonchi, Arijit Khan, and André Panisson}

\begin{abstract}
Correlation Clustering (CC) is a natural formulation of clustering in combinatorial optimization, which uses a graph representation of the input and does not require a pre-specified number of clusters. Given $n$ objects and a pairwise similarity function, the
goal is to cluster the objects so that
similar objects are put in the same cluster and dissimilar objects
are put in different clusters. Despite its versatility, existing CC algorithms suffer from significant scalability issues and are inherently transductive: i.e., the algorithm must be executed from scratch for any new problem instance.

In this work, we bridge this gap by leveraging Graph Neural Networks (GNNs) to solve Inductive Correlation Clustering, a novel generalization of the CC problem designed to handle unseen graph instances. By learning to exploit common structural patterns and node features during training, our framework generalizes to new graphs drawn from the same distribution with minimal computational overhead with respect to standard algorithms. We demonstrate the effectiveness and scalability of our approach through extensive experiments. Our framework not only excels in the inductive setting, e.g., lowering the inference time up to $5$ order of magnitude, while maintaining an approximation ratio within $~10\%$ of the best baseline solution,
but also achieves competitive results on standard (transductive) CC benchmarks. Finally, we showcase a practical application of our framework as a learnable pooling mechanism for graph classification. Our results indicate that our method serves as an efficient pooling layer, enhancing the ability of GNNs to capture hierarchical structural information in networks.
\end{abstract} 


%

\maketitle \sloppy 

\section{Introduction}

Clustering is a foundational data analysis task aimed at grouping similar objects while separating dissimilar ones. A natural formulation of such objective is provided, in graph terminology, by Correlation Clustering (CC) \cite{bansal2004correlation,bonchi2022correlation}: given an unweighted graph $G=(V,E)$, where the set of nodes $V$ represents the objects to be clustered and edges in $E$ connect \emph{similar} objects,
the goal is to partition $V$ into clusters so to minimize the total number of \emph{``disagreements''}, defined as the sum of missing edges within clusters and existing edges between clusters. Such optimization problem naturally extends to non-binary, symmetric similarity functions, and to settings where the similarity information is incomplete, i.e., no penalty needs to be paid for some pairs of objects irrespective of whether they are placed together or not. A key feature of CC is that,
in contrast to many other popular clustering techniques (such as $k$-means),
it does not require the number of clusters to be given as input, instead the optimal number of clusters is naturally found by the process. Since it only requires as input a notion of similarity, CC is particularly appealing for the task of clustering structured objects, where the similarity function is domain-specific and can be designed or learned. Thanks to its broad generality, CC has been studied under many variants (see, e.g., \cite{cc_tutorial,bonchi2022correlation}), and in a multitude of downstream applications ranging from duplicate detection and similarity joins~\cite{duplicate_detection,kushagra2019semi,demaine2006correlation},
to biology~\cite{dasgupta07algorithmic,clustering_genes} and computer vision, specifically to the problem of \emph{image segmentation}~\cite{bjorn11probabilistic,image_segmentation,yarkony12fast}.

However, CC faces significant challenges. Its general formulation is APX-hard \cite{charikar2005clustering,demaine2006correlation}, and existing approximation algorithms often scale poorly. Many rely on linear programs with exhaustive constraint sets, making them infeasible for large graphs. Furthermore, traditional CC algorithms typically ignore node features, overlooking rich information that could better identify latent structures within the graph. Finally,
traditional CC is inherently transductive, requiring to be re-run for each new problem instance, without the possibility of generalizing to entirely new instances.

\vspace{-1mm}
\spara{Contributions.} In this paper, we introduce \emph{Inductive Correlation Clustering}, a new formulation of CC which can generalize to unseen instances. Hierarchical pooling, entity resolution, and subgraph detection are examples of tasks that can benefit from inductive clustering models. In hierarchical pooling, graph data must be compressed into meaningful ``super-nodes'', for example, grouping atoms into functional units for molecular property prediction \cite{Wang2024}, or segment point clouds for computer vision tasks \cite{Chen_2016_CVPR}. An entity resolution task, on the other hand, resolve identities across databases through relational clustering, which is vital for many dataset deduplication applications \cite{entityres2021Christophides}. Subgraph detection can instead help identifying suspicious communities in transaction monitoring \cite{starnini2021smurf,yow2023machine}, or groups on social networks \cite{gao2021ics}. In all these scenarios, graphs are processed in batches or continuous streams during both training and inference. Solving a CC instance from scratch for every unique graph in a batch is computationally prohibitive compared to inductive methodologies, especially for larger graphs. By adopting an inductive, scalable approach based on CC, we can perform these tasks even with very little information regarding the types of relationships across elements or the number of underlying clusters, while providing the speed and generalization required for large-scale or real-time deployments on multiple graphs.

Graph Neural Networks (GNNs) \cite{scarselli2005graph,kipf2017semisupervised} have recently emerged as effective tools for tackling combinatorial optimization problems \cite{schuetz2022combinatorial,cappart2023combinatorial}. Their key strength lies in generalizing across diverse problem instances \cite{joshi2022learning,Nerem2025shortest} by capturing structural patterns and exploiting node-level features. This makes GNN-based approaches strong candidates for Inductive Correlation Clustering.

We thus propose a framework to train GNNs for solving Inductive Correlation Clustering problems. By employing a loss function based on a differentiable relaxation of the standard Correlation Clustering objective, our framework can learn to optimize via stochastic gradient descent in an unsupervised manner, allowing for an efficient training and inference. To address the fundamental differences between transductive and inductive settings, our framework introduces two distinct architectural variants. The inductive variant, once trained, generalizes to previously unseen graph instances, producing solutions of comparable quality to traditional algorithms while being orders of magnitude faster. We demonstrate the utility of Inductive Correlation Clustering by applying it to graph pooling. As mentioned before, pooling is a critical technique for deep learning, allowing models to capture high-level structural information, increase the receptive field, and reduce computational overhead by pruning edges and coarsening the graph \cite{wang2024graph}. Our experiments indicate that our methodology serves as a highly effective pooling technique, competitive with state-of-the-art approaches.

In summary, we make the following contributions:

\begin{squishlist}
    \item \textbf{Novel problem formulation:} We introduce the Inductive Correlation Clustering problem, a generalization of Correlation Clustering that leverages graph structure and node features to infer clustering solutions on unseen graphs drawn from the same distribution (\S \ref{sec:prob_statement}).
    \item \textbf{Scalable framework:} We propose a scalable GNN-based training framework designed specifically to solve Correlation Clustering instances, which we dub \textsf{CC-GNN}, featuring a transductive (\textsf{NodeGNN}) and an inductive (\textsf{LinkGNN}) variant (\S \ref{sec:framework}).
    \item \textbf{Empirical validation:} We demonstrate that \textsf{LinkGNN} provides high-quality and computationally efficient solutions for the inductive setting, while \textsf{{NodeGNN}} excels on standard Correlation Clustering instances (\S \ref{sec:exp_setup}-\S \ref{sec:exp}).
    \item \textbf{Downstream application:} We showcase the practical utility of our framework as a learnable pooling layer for deep graph learning architectures (\S \ref{sec:pooling_exp}).
\end{squishlist}

\noindent Our work, bridging two apparently remote areas, such as \emph{combinatorial optimization for clustering} and \emph{deep learning on graphs}, contributes to the growing body of literature showing the successful usage of GNNs in combinatorial optimization problems \cite{schuetz2022combinatorial,cappart2023combinatorial}. 

\vspace{-2mm}
\section{Related Work}
\vspace{-1mm}
\spara{Correlation clustering.} Correlation Clustering is a classical optimization problem with a rich theoretical history \cite{bansal2004correlation, charikar2005clustering, bonchi2022correlation}. Much of the existing literature focuses on the theoretical development of approximation algorithms \cite{cao2025solving, cohen2021correlation, chakrabarty2023single}. However, implementations capable of providing high-quality approximations while scaling to large datasets remain scarce or are limited to simplified cases \cite{veldt2022correlation, bengali2023faster, behnezhad2025correlation}. One of the seminal works introduced KwikCluster  \cite{ailon2008aggregating}, a combinatorial algorithm providing a $3$-approximation guarantee in expectation for unweighted, complete graphs. More recently, \cite{behnezhad2025correlation} proposed ModifiedPivot, which improves this bound to $3-\epsilon_0$ for some $\epsilon_0 > 0$ on the same type of graphs. Both algorithms have been shown to maintain their approximation guarantees in dynamic settings, where the graph undergoes edge insertions or deletions~\cite{behnezhad2019fully,dalirrooyfard2024pruned}.
A complementary line of research \cite{charikar2005clustering,demaine2006correlation} tackles CC by solving a Linear Programming (LP) relaxation followed by a rounding procedure. While these methods often yield tighter approximation bounds \cite{cohen2024combinatorial,cao2025solving}, they need solving LPs with an extremely large number of constraints. To apply these techniques to real-world graphs, additional approximations are required to maintain computational feasibility \cite{veldt2022correlation,bengali2023faster}.

Online \cite{cohen2022online,mathieu2010online}, active \cite{Garcia-SorianoK20,Kuroki0B024} streaming \cite{pmlr-v37-ahn15,behnezhad2023single,chakrabarty2023single}, and dynamic \cite{behnezhad2019fully,behnezhad2025correlation,dalirrooyfard2024pruned} variants of CC have been studied extensively: these variants typically
 assume a continuous flow of node/edge modifications to a single problem instance. In contrast, our work focuses on the inductive setting: learning from a distribution of graphs to cluster entirely new, previously unseen instances with minimal latency.

\spara{Inductive clustering on graphs.}
Clustering methodologies can be categorized into transductive (i.e., must be run from scratch on unseen problem instances) and inductive approaches (i.e., able to learn a generalizable function that can be applied to unseen problem instances) \cite{miyamoto2012inductive}.
Within the domain of graph-structured data, most standard clustering techniques are transductive, e.g., Leiden algorithm \cite{traag2019louvain}, spectral clustering methods \cite{nascimento2011}, and standard Correlation Clustering algorithms. Inductive clustering methods leverage node features to generalize across instances \cite{Lun2022inductive}. The most significant methods in this area use GNNs \cite{scarselli2005graph,kipf2017semisupervised}, with the shift toward the inductive setting popularized by GraphSAGE \cite{hamilton2017inductive}.
Notable examples include DMoN~\cite{tsitsulin2023graph}, HOSC~\cite{duval2022higher}, GTV~\cite{hansen2023total}, and MinCutPool~\cite{bianchi2020spectral}. While all of these methods require setting a fixed or maximum number of clusters, our Inductive Correlation Clustering formulation allows to select the number of clusters according to the structure and density of the input graph.

\spara{Graph pooling.} A primary application of inductive approaches is graph pooling \cite{wang2024graph,gal2025towards}, where learnable clustering serves to coarsen graphs. This process not only manages computational complexity but also enhances the expressivity of the resulting GNN architecture \cite{bianchi2023expressive}. We compare with many of these methods in a pooling task in \S \ref{sec:pooling_exp}. Besides, we also compare against $k$-MIS~\cite{bacciu2023generalizing}, MaxCutPool~\cite{abatemaxcutpool}, and SEP~\cite{wu2022structural} which allows to not set a fixed number of output nodes. However, these are not clustering techniques, since they coarsen graphs by selecting subsets of representative nodes instead of identifying communities; $k$-MIS is not even inductive, as it must run separately on each new graph and does not use node features. 

\section{Inductive Correlation Clustering}\label{sec:prob_statement}
Consider an unweighted graph $G=(V, E)$.
The objective of Correlation Clustering is to partition the set of nodes $V$ into clusters that minimize the total number of ``disagreements'', defined as the sum of missing edges within clusters and existing edges between clusters.
Let $\mathbf{C}\in\{0,1\}^{|V|\times K}$ be a cluster assignment matrix, where each node is assigned to exactly one of $K$ clusters ($K$ is not fixed a priori). The cost of a clustering $\mathbf{C}$ is formalized as:
\begin{equation}
\label{eq:node_wise_formulation}
    \text{cost}(\mathbf{C})= \sum_{(i,j)\in E}\sum_{c=1}^K \mathbf{C}_{ic} (1-\mathbf{C}_{jc} ) + \sum_{(i,j)\not\in E}\sum_{c=1}^K \mathbf{C}_{ic}\mathbf{C}_{jc}.
\end{equation}
This objective can be expressed more intuitively by introducing the co-clustering matrix $\mathbf{M}=\mathbf{C}\mathbf{C}^\intercal$, where $\mathbf{M}_{i,j}=1$ if nodes $i$ and $j$ are in the same cluster, and $0$ otherwise.
Therefore, the cost function simplifies to:
\begin{equation}
\label{eq:link_wise_formulation}
    \text{cost}_{G}(\mathbf{M})= \sum_{(i,j)\in E}\left(1-\mathbf{M}_{i,j}\right) + \sum_{(i,j)\not\in E}\mathbf{M}_{i,j}.
\end{equation}
Minimizing Eq.~\eqref{eq:link_wise_formulation} is known to be APX-hard~\cite{bansal2004correlation}.
To address this, the problem is often relaxed by allowing $\mathbf{M}_{ij}\in[0,1]$ subject to metric constraints (i.e., triangle inequality $1-\mathbf{M}_{ij} \le (1-\mathbf{M}_{ik}) + (1-\mathbf{M}_{jk})$, and symmetry $\mathbf{M}_{ij}=\mathbf{M}_{ji}$).
This relaxation forms the basis of Linear Programming approaches~\cite{charikar2005clustering, bengali2023faster} and motivates the continuous relaxations we adopt for our GNN framework.

In the inductive setting, we assume that graphs are drawn from a distribution $\mathcal{D}$. A graph instance is defined as a tuple $G=(V, E, \mathbf{X})$, where $\mathbf{X} \in \mathbb{R}^{|V|\times F}$ represents node features.
We aim to learn a parameterized function $f_\theta$ (e.g., a GNN) that maps a graph instance to a clustering assignment.
Given a training graph $G_{tr}$ (or a set of graphs), the goal is to find the parameters $\theta^*$ that minimize the clustering cost on the training structure:
\begin{equation}
    \theta^*=\arg\min_{\theta} \text{cost}_{G_{tr}}\left(f_{\theta}(V_{tr}, E_{tr}, \mathbf{X}_{tr})\right).
\end{equation}
The ultimate objective is to generalize this solution to minimize the cost on a disjoint, previously unseen test graph $G_{te} \sim \mathcal{D}$:
\begin{equation}
    \text{Objective: } \min \;\text{cost}_{G_{te}}\left(f_{\theta^*}(V_{te}, E_{te}, \mathbf{X}_{te})\right).
\end{equation}

We show in the following sections how we learn the model $f_{\theta}$, and demonstrate via experiments the effectiveness of our approach. 

\section{CC-GNN Framework}
\label{sec:framework}
In this section, we present \textsf{CC-GNN}, a framework for Correlation Clustering in transductive and inductive settings. We distinguish between two optimization approaches: node-wise methods, which optimize a relaxed cluster assignment matrix $\mathbf{C} \in [0,1]^{|V| \times K}$, and edge-wise methods, which optimize the relaxed co-clustering matrix $\mathbf{M} \in [0,1]^{|V| \times |V|}$.

Formally, a GNN $f_\theta$ maps node features $\mathbf{X}$ and the edge set $E$ to a latent embedding matrix $\mathbf{O} \in \mathbb{R}^{|V| \times F}$:
\begin{equation}
    \mathbf{O} = \text{GNN}_{\theta}(\mathbf{X}, E).
\end{equation}

\noindent The initialization of $\mathbf{X}$ defines the model's operating mode. In transductive settings lacking informative attributes, we follow \cite{sato2021random, abboud2020surprising} and use random features as unique identifiers to enhance expressivity. While structural features (e.g., Laplacian eigenvectors) are an alternative, they offered no empirical advantage in our preliminary experiments. Conversely, for inductive generalization to unseen graphs, we utilize node attributes to capture transferable patterns.


The framework is organized into three distinct components. First, we introduce a \textit{pivot-based sampling} strategy that ensures scalability to large graphs (\S \ref{sec:pivot}). Second, we describe a \textit{node-wise optimization} approach for standard, transductive Correlation Clustering (\S \ref{sec:transductive}). Finally, we present our primary contribution: an \textit{inductive, link-wise formulation} that enables the model to infer clustering solutions on entirely new graph instances without re-training (\S \ref{sec:inductive_intuition}-\S\ref{sec:inductive_method}).


\subsection{Pivot-based Batch Sampling}
\label{sec:pivot}
Since message-passing operations scale poorly on large graphs, optimizing over the full training graph $G_{tr}$ is often infeasible. Instead, we perform backpropagation on a subgraph induced by a node batch $B\subseteq V_{tr}$. As detailed in Algorithm \ref{alg:pivot_sampling}, we utilize a sampling strategy inspired by KwikCluster \cite{ailon2008aggregating}: at each epoch we select $M$ random pivots $P$ and include all their direct neighbors to form $\mathcal{B}$.

In the original KwikCluster algorithm, each pivot and its neighbors are iteratively extracted to form clusters. By adopting this logic during training, we identify potential clusters and "refine" them by optimizing the model loss on these localized communities rather than the entire graph. This approach significantly reduces computational overhead, lowering the complexity of each GNN call from $O(E)$ to $O(\frac{M}{|V_{tr}|}E)$. This provides a substantial performance advantage when $M \ll |V_{tr}|$. As we will see in \S \ref{sec:ablation_study}, optimizing over a batch also drastically reduces the computational complexity of the loss computation, reducing training time by up to a factor of $40$ and memory consumption by a factor of $4$. 

\begin{algorithm}[t!]
\caption{Pivot-based Sampling}
\label{alg:pivot_sampling}
\begin{algorithmic}[1]
\REQUIRE Graph $G = (X, E)$, signed adjacency $\mathbf{W}$, \# pivots $M$
\ENSURE Batched features $\mathbf{X}_B$, edges $E_B$, weight matrix $\mathbf{W}_B$

\STATE Select a set of $M$ random pivots $P \subset V$
\STATE $B \gets P$
\FOR{each $v \in P$}
    \STATE $B \gets B \cup \mathcal{N}(v)$
\ENDFOR

\STATE $E_B = \{(u,v) \in E \mid u,v \in B\}$
\STATE $\mathbf{X}_B = \{\mathbf{X}_u \mid u \in B\}$
\STATE $\mathbf{W}_B = \mathbf{W}[B, B]$ $\backslash^*$ Submatrix induced by node set $B$ $^*\backslash $
\RETURN $\mathbf{X}_B ,E_B, \mathbf{W}_B$

\end{algorithmic}
\end{algorithm}

\begin{algorithm}[t!]
\caption{NodeGNN Training}
\label{alg:nodewise_training}
\begin{algorithmic}[1]
\REQUIRE Graph $G = (X, E)$, signed adjacency $\mathbf{W}$, epochs $N$, maximum number of clusters $K$, pivots number $M$.
\ENSURE Optimal GNN parameters $\theta^*$, cluster assignments $\mathcal{C}$
\STATE Initialize GNN parameters $\theta$

\FOR{epoch = 1 \TO $N$}
    \STATE  $\mathbf{X}_B, E_B, \mathbf{W}_B$ $\leftarrow \text{PivotSampling}(G, \mathbf{W}, M)$ $\backslash^*$ Call Alg. \ref{alg:pivot_sampling} $^*\backslash$
    \STATE $\mathbf{O} = \text{GNN}_{\theta}(X_B, E_B)$

    \STATE  $\mathbf{C}_{i} = \text{softmax}(\mathbf{O}_i)$

    \STATE Calculate loss: $\mathcal{L}(\mathbf{C}) = \lVert \mathbf{W-CC}^{\intercal}\rVert_2^2 - \lVert\mathbf{CC}^{\intercal}\rVert_2^2$

    \STATE Backpropagate $\nabla_{\theta} \mathcal{L}$ and update $\theta$
\ENDFOR
\STATE $\theta^* \gets \theta$
\STATE Generate final embeddings $\mathbf{O}^* = \text{GNN}_{\theta^*}(X, E)$

\FOR{each node $i \in V$}
    \STATE $\mathcal{C}_i = \text{argmax}_{j \in \{1, \dots, K\}} \mathbf{O}^*_{ij}$
\ENDFOR

\RETURN $\theta^*, \mathcal{C}$

\end{algorithmic}
\end{algorithm}

\subsection{Transductive Node-wise Optimization}
\label{sec:transductive}

To train GNNs for Correlation Clustering, we relax the cost function into a continuous, differentiable loss. A key advantage of deep learning here is the ability to satisfy constraints implicitly through model architecture rather than explicit enforcement. This makes local optimization feasible even for large graphs where the number of constraints would otherwise be prohibitive. We focus first on the node-wise approach, and call this method \textsf{NodeGNN}; its training procedure is detailed below and summarized in Algorithm \ref{alg:nodewise_training}.

In order to optimize Eq. \ref{eq:node_wise_formulation}, we first set the value of $K$ arbitrarily and relax the node assignment matrix as $\mathbf{C} \in [0, 1]^{|V| \times K}$.
This restricts the solution space to partitions with at most $K$ clusters, a constraint absent from the original formulation with strong repercussions, as we will see.

In this setting, the cluster assignment matrix relaxation has the constraints $\sum_j \mathbf{C}_{ij}=1$, $\forall i \in V$. If we set the latent embeddings dimension $F$ equal to $K$, these constraints are easily satisfied if we define $\mathbf{C}=\text{softmax}(\mathbf{O})$, with $\mathbf{O}$ the node embedding matrix output of a GNN (lines 5-6 of Algorithm \ref{alg:nodewise_training}).

To construct a differentiable objective, we first define the signed adjacency matrix $\mathbf{W}$ as:
\begin{equation}
\mathbf{W}_{ij}=
    \begin{cases}
        +1, & \text{if}\; i\neq j\; \text{and}\; (i,j)\in E \\
        -1,&\text{if}\; i\neq j\; \text{and}\; (i,j) \not\in E\\
        0, & \text{if} \; i=j.
    \end{cases}
\end{equation}
Introducing $\mathbf{W}$ in Eq. \ref{eq:node_wise_formulation} and simplifying we can reduce it to:
\begin{align}
    \text{cost}_{G}(\mathbf{C})&= -\frac{1}{2}\sum_{ij}\mathbf{W}_{ij}\sum_c\mathbf{C}_{ic}\mathbf{C}_{jc} +|E|\\  &= -\frac{1}{2}\text{Trace}(\mathbf{WCC}^\intercal)+\text{const}.
\end{align}
Exploiting the expansion of the squared Frobenius norm (the entry-wise 2-norm), we can also cast the trace as:
\begin{equation}
\label{eq:matrix_formulation}
    -\text{Trace}(\mathbf{WCC}^\intercal) = \frac{1}{2}\left( \lVert \mathbf{W}- \mathbf{CC}^{\intercal}\rVert_2^2 -\lVert\mathbf{CC}^{\intercal}\rVert_2^2 - \lVert \mathbf{W}\rVert_2^2\right).
\end{equation}
Discarding the constant term $\lVert \mathbf{W}\rVert_2^2$, we arrive at the final loss function, which favours the alignment between $\mathbf{W}$ and $\mathbf{C}\mathbf{C}^\top$ :
\begin{equation}
\label{eq:node_wise-loss}
    \mathcal{L}(\mathbf{C}) =\lVert \mathbf{W}- \mathbf{CC}^{\intercal}\rVert_2^2 - \lVert\mathbf{CC}^{\intercal}\rVert_2^2.
\end{equation}
Computing this loss over the full graph has time complexity of $O(|V|^2K)$. Consequently, a regime where $K \ll |V|$ significantly reduces this overhead, though at the expense of restricting the reachable solutions.

At inference time, instead, we assign clusters through an argmax on the output $\mathbf{O}^*$ from the GNN (lines 12-14 of Algorithm \ref{alg:nodewise_training}).

Both the trace and matrix formulations facilitate local optimization of $\mathbf{C}$;
however, Eq. \ref{eq:matrix_formulation} 
specifically frames Correlation Clustering as a matrix factorization problem. The target matrix $\mathbf{W}$ is approximated by $\mathbf{CC}^{\intercal}$, with $\lVert\mathbf{CC^{\intercal}}\rVert_2^2$serving as a regularization term. GNN architectures are well suited here, as $\mathbf{W}$ and  the rows of $\mathbf{C}$ can naturally correspond to a weighted signed graph and $K$-dimensional node embeddings $\mathbf{O}$, respectively. By leveraging the graph structure to learn these node embeddings, GNNs effectively perform the factorization required to optimize the clustering objective~\cite{zhang2019inductive, huang2021mixgcf}.

Since this approach stems from fixing the maximum number of cluster $K$ and optimizing the corresponding matrix $\mathbf{C}$, as the graph size increases, the approximation quality drops significantly unless $K$ is scaled accordingly. Additionally, this matrix factorization approach is strictly transductive. Because the GNN learns to factorize the specific matrix $\mathbf{W}$ of the training graph, it fails to generalize to unseen data. Even when applied to new graphs with node features drawn from the same distribution, it cannot transfer its learned representations and outputs incoherent clusters (see \S \ref{sec:iCC_exp}). To address these challenges, the following sections introduce an inductive, edge-wise optimization approach. 

\subsection{Intuition on Inductive Optimization}
\label{sec:inductive_intuition}

To overcome the inherent limitations of the node-wise approach, we consider a link-wise optimization approach, i.e., we optimize with respect to the co-clustering matrix $\mathbf{M}\in[0,1]^{|V|\times|V|}$.

\begin{figure}[t!]
\centering
\includegraphics[width=1\linewidth]{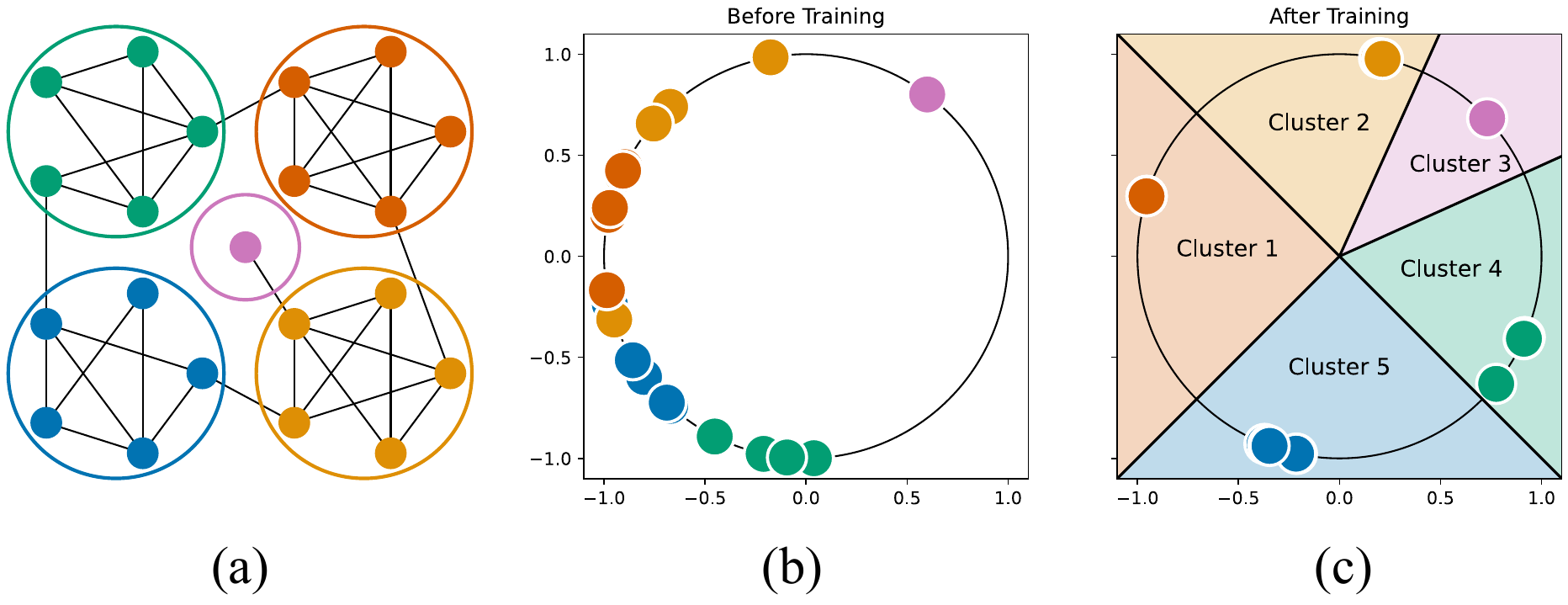}
\vspace{-5mm}
    \caption{
(a) A toy graph instance where the optimal CC solution is highlighted by colored circles (ground truth).
The corresponding node embeddings generated by \textsf{LinkGNN} are projected onto the unit circle ($F=2$). Before training (b), embeddings are randomly distributed. After training (c), the model separates the nodes into 5 distinct angular sectors.
\label{fig:example}}
\vspace{-5mm}
\end{figure}

Before detailing the method, which we call \textsf{LinkGNN}, we illustrate our intuition with an example. Consider the graph in Figure \ref{fig:example}a, where the optimal Correlation Clustering solution consists of four small communities (5 nodes each) and a single isolated node. To enable visualization, we use a GNN with an output dimension of $F=2$ and project the resulting node embeddings onto a circle to constrain possible distances (in general applications, this is extended to a hypersphere of dimension $F$). The distances between these projected embeddings then define the relaxed co-clustering matrix $\mathbf{M}$, as will be described in the following section.

Our loss acts similarly to a contrastive learning objective. Initially, a GNN produces node embeddings that are clustered tightly together (Figure \ref{fig:example}b). Through training, the loss pulls connected nodes closer and pushes disconnected ones apart until a local optimum is reached. In our example, this process projects nodes into clearly identifiable clusters along the unit circle (Figure \ref{fig:example}c). Post-training, we retrieve the optimal solution by grouping nodes based on their proximity in the latent space. Because clustering depends on these distances rather than fixed assignments, the model is not restricted to a maximum number of clusters ($K$), allowing it to adapt \textit{inductively} to graphs of varying scales.

\subsection{Inductive Link-wise Optimization}
\label{sec:inductive_method}

In \textsf{LinkGNN}, we optimize a relaxed link-formulation of CC. 
When solving the link-wise formulation, we need a relaxation $\mathbf{M}$ with the metric constraints described in \S \ref{sec:prob_statement}. To do so, we first normalize the node embeddings:
\begin{equation}
\label{eq:normalization}
    \mathbf{\tilde{O}}_i = \frac{\mathbf{O}_i}{\lVert \mathbf{O}_i\rVert_2}, ~~\forall  \, i \in V.
\end{equation}
We can then obtain matrix $\mathbf{M}$ by computing:
\begin{equation}
    \mathbf{M}_{ij} = 1- \frac{1}{2}\lVert\mathbf{\tilde{O}}_i  -\mathbf{\tilde{O}}_j\rVert_2,
\end{equation} 
where we use $\lVert\cdot\rVert$ to indicate the Euclidean norm. It is necessary to do this step both during training (lines 5-7 of Algorithm \ref{alg:link_wise_training}) and inference (lines 1-3 of Algorithm \ref{alg:inductive_inference}).

Thanks to the properties of the Euclidean norm, the matrix $\mathbf{M}$ is now symmetric ($\mathbf{M}_{ij}=\mathbf{M}_{ji}$) and satisfies the triangular inequalities $1-\mathbf{M}_{ij} \le (1-\mathbf{M}_{ik}) + (1-\mathbf{M}_{jk})$ for all $i,j,k$, as required by the constraint in \S \ref{sec:prob_statement}.
However, by normalizing the node embeddings $\mathbf{O}$, we are effectively limiting the matrix $\mathbf{M}$ to describe distances between points on an $F$ dimensional hypersphere of unit radius. This is an additional constraint on the possible values of the matrix $\mathbf{M}$ with respect to the classical relaxation of \cite{charikar2005clustering, demaine2006correlation}. We can then cast the loss function in Eq. \ref{eq:node_wise-loss} in terms of the co-clustering matrix $\mathbf{M}$ obtaining:
\begin{equation}
    \label{eq:edge_wise-loss}
    \mathcal{L}(\mathbf{M}) =\lVert \mathbf{W}- \mathbf{M}\rVert_2^2 - \lVert\mathbf{M}\rVert_2^2.
\end{equation}
The loss still requires the computation of $\mathbf{M}$, which has a computational complexity of $O(|V|^2F)$ on the full graph: however, since we are using the pivot-based sampling, we are able to reduce it to $O(|\mathcal{B}|^2F)$. Since we can also set the latent dimension $F$ to a much smaller value than previous maximum number of clusters $K$, it is also significantly faster than the node-wise loss.

During inference, we obtain the clusters by applying another heuristic approach (Algorithm \ref{alg:inductive_inference}), where we define the matrix:
\begin{equation}
\label{eq:threshold_graph}
\vspace{-2mm}
    \mathbf{\hat{M}}^{t}_{ij}=
    \begin{cases}
        1, \; &\text{if}\; (i,j)\in E \; \text{and} \; \mathbf{M}_{ij}\geq t \\
        0, \; &\text{otherwise}.
    \end{cases}
\vspace{-0.1mm}
\end{equation}
Since the matrix $\mathbf{\hat{M}}^{t}$ acts as an adjacency matrix derived from the similarity matrix $\mathbf{M}$, we define the output clusters as its connected components. Finally, because the threshold $t$ is arbitrary, we evaluate a range of values $t \in [0, 1]$ during training and select the one that minimizes the cost (Algorithm \ref{alg:link_wise_training}, lines 11-18). The corresponding $t$ is then used for inference as well. Increasing $t$ sparsifies the graph, leading to a higher number of output clusters as more connected components are formed. After training, we can then use this heuristic to cluster the original training graph (as we do when we limit to a transductive application of the method) or to new graphs entirely (allowing for inductive applications). The main drawback is that this heuristics, together with our relaxation, can introduce a larger approximation error than \textsf{NodeGNN} (as we observe experimentally in \S \ref{sec:gurobi_Vs_gnn} and \S \ref{sec:tCC_exp}).

Since during inference we only need to compute distances $\mathbf{M}_{ij}$ for existing edges, we significantly reduce the auxiliary graph's density and greatly accelerate the inference (reducing the computation of $\mathbf{M}$ from $O(|V|^2F)$ to $O(|E|F)$). Focusing on already connected nodes can also be theoretically justified by the following observation: if a cluster contains two communities not linked by any edge, separating them yield a lower cost than keeping them merged. Accounting for the matrix multiplications in the message passing operations, inference time for LinkGNN scales as $\mathcal{O}(|V|F^2 + |E|F)$. Furthermore, inference memory consumption scales as $\mathcal{O}(F^2 + |V|F + |E|)$. 

Initializing nodes with random feature vectors can be effective for transductive tasks, but it inherently limits generalization to unseen graphs. In such cases, the GNN memorizes node identities from the training set rather than learning transferable structural patterns. Consequently, with \textsf{LinkGNN}, node features (when available) or structure-based embeddings such as Node2Vec must be used. For pure Correlation Clustering objectives, Node2Vec embeddings are often preferable to raw node features, since original features may be entirely uncorrelated with graph structure, the sole determinant of the optimal solution.

\begin{algorithm}[t!]
\caption{LinkGNN Training}
\label{alg:link_wise_training}
\begin{algorithmic}[1]
\REQUIRE Graph $G = (X, E)$, signed adjacency $\mathbf{W}$, epochs $N$, number of pivots $M$, lists of thresholds $T$
\ENSURE Trained GNN parameters $\theta^*$, optimal threshold $t^*$
\STATE Initialize GNN parameters $\theta$
\FOR{epoch = 1 \TO $E_{max}$}
    \STATE  $\mathbf{X}_B,E_B,\mathbf{W}_B \leftarrow \text{PivotSampling}(G, \mathbf{W}, M)$ $\backslash^*$ Call Alg. \ref{alg:pivot_sampling} $^*\backslash$
    \STATE $\mathbf{O} = \text{GNN}_{\theta}(X_B, E_B)$
    \STATE $\mathbf{\tilde{O}}_i = \mathbf{O}_i / \lVert \mathbf{O}_i\rVert_2$ $\backslash^*$ Row-wise Normalization $^*\backslash$
    \STATE $\mathbf{M}_{ij} = 1 - \frac{1}{2}\lVert\mathbf{\tilde{O}}_i - \mathbf{\tilde{O}}_j\rVert_2$
    \STATE $\mathcal{L}(\mathbf{M}) = \lVert \mathbf{W}_B - \mathbf{M}\rVert_2^2 - \lVert\mathbf{M}\rVert_2^2$
    \STATE Update $\theta$ via backpropagation of $\nabla_{\theta} \mathcal{L}$
\ENDFOR
\STATE $\backslash^*$Best threshold selection$^*\backslash$
\STATE $MinCost \gets \infty$, $t^* \gets 0$
\FOR{each $t \in T$}
    \STATE $\mathcal{C}_t \leftarrow \text{Inference}(G, \text{GNN}_{\theta^*}, t)$ $\backslash^*$ Call Alg. \ref{alg:inductive_inference} $^*\backslash$
    \STATE $CurrCost = \text{Cost}_{G}(\mathcal{C}_t)$ $\backslash^*$ Evaluate cost with eq. \ref{eq:node_wise_formulation} $^*\backslash$
    \IF{$CurrCost < MinCost$}
        \STATE $MinCost \gets CurrCost$, $t^* \gets t$
    \ENDIF
\ENDFOR
\RETURN $\theta^*, t^*$
\end{algorithmic}
\end{algorithm}

\begin{algorithm}[t!]
\caption{LinkGNN Inference}
\label{alg:inductive_inference}
\begin{algorithmic}[1]
\REQUIRE Graph $G = (X, E)$, trained GNN, threshold $t$
\ENSURE Cluster assignments $\mathcal{C}_t$

\STATE $\mathbf{O} = \text{GNN}(X, E)$
\STATE $\mathbf{\tilde{O}}_i = \mathbf{O}_i / \lVert \mathbf{O}_i\rVert_2$ $\backslash^*$ Row-wise Normalization $^*\backslash$
\STATE $\mathbf{M}_{ij} = 1 - \frac{1}{2}\lVert\mathbf{\tilde{O}}_i - \mathbf{\tilde{O}}_j\rVert_2$

\STATE $\mathbf{\hat{M}}^{t}_{ij} =
\begin{cases}
1, & \text{if } (i,j) \in E \text{ and } \mathbf{M}_{ij} \geq t \\
0, & \text{otherwise}
\end{cases}$

\STATE $\mathcal{C}_t \gets \text{ConnectedComponents}(\mathbf{\hat{M}}^{t})$
\RETURN $\mathcal{C}_t$
\end{algorithmic}
\end{algorithm}

\subsection{Pooling Adaptation}
\label{sec:pooling}
Pooling applications require to modify our method slightly. We consider only the link-wise optimization approach, motivated by the experimental results on inductive applications (\S \ref{sec:iCC_exp}).

\begin{figure}[t]
    \vspace{-3mm}
\begin{tabular}{cc}
\includegraphics[width=.3\linewidth]{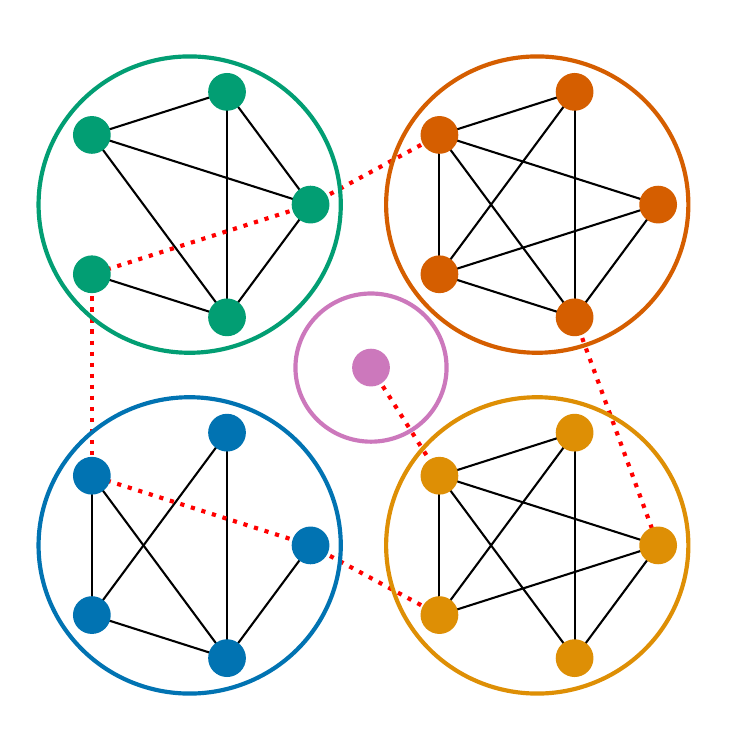} &
\includegraphics[width=.3\linewidth]{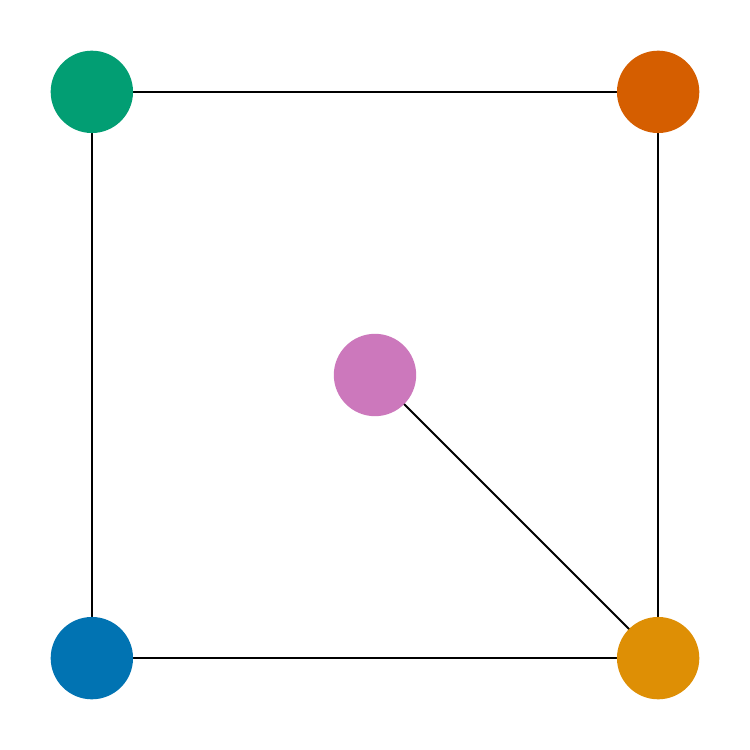}\\
(a) & (b) 
\end{tabular}
    \vspace{-3mm}
    \caption{Example of pooling on the graph of Figure \ref{fig:example} with $r=20\%$. (a) Red dotted lines indicate the $20\%$ of edges connecting node pairs with the greatest embedding distance. (b) The pooling operation merges nodes within the same connected components and then re-introduce their original connections: the graph is significantly compressed, retaining fewer than $15\%$ of original edges.
    }
    \label{fig:run_example_pooling}
    \vspace{-6mm}
\end{figure}

While originally we optimize only the CC cost, and select the threshold $t$ accordingly, in pooling we must always be able to regulate the coarsening of the graph through an hyperparameter. Therefore, we introduce an \textit{edge pool percentile} parameter $r \in [0, 100]$. We dynamically set the threshold $t$ to the $(100-r)$-th percentile of the entries in the similarity matrix $\mathbf{M}$.
This ensures that we retain only the strongest connections, effectively controlling the coarsening ratio. After inferring the clusters from $\mathbf{M}$, we first obtain the coarsened nodes representations by performing a sum aggregation across the nodes in each individual clusters. Then, we link any two of the clusters if their constituent nodes shared at least one edge in the original graph. These choices on aggregation and link reconstruction are in line with most pooling frameworks \cite{grattarola2022understanding}. Finally, because optimal pooling depends on both the CC objective and the downstream task loss, we use the node representations from the downstream model as input embeddings instead of the Node2Vec embeddings of LinkGNN. These embeddings are initialized directly from the original node features before being updated by the downstream model. The CC loss is also \textit{not} backpropagated to the downstream model. We present an example in Figure \ref{fig:run_example_pooling}.

\section{Experimental Setup}
\label{sec:exp_setup}

To rigorously assess the proposed \textsf{CC-GNN} framework, we construct a unified experimental suite, targeting 4 research questions:

\sparas{RQ1 (Approximation quality):} Does the proposed differentiable relaxation enable the GNN to recover high-quality solutions relative to exact solvers on tractable instances?

\sparas{RQ2 (Transductive performance):} How does our framework compare against state-of-the-art combinatorial algorithms on standard Correlation Clustering benchmarks without node features?

\sparas{RQ3 (Inductive generalization):} Can the model exploit structural patterns and node features to generalize to unseen graphs without retraining, and does this yield inference-time efficiency gains?

\sparas{RQ4 (Downstream utility):} Is Inductive Correlation Clustering effective when used as a learnable pooling layer (\textsf{CCPool}) within deep graph learning architectures?

\smallskip

\noindent\textbf{Reproducibility:} Our code and data are available at \url{https://github.com/FrappaN/ccgnn}.

\subsection{Datasets}

We use a diverse collection of graph datasets categorized into three groups.
Summary statistics for all datasets are reported in Table~\ref{tab:unified_datasets}.

\spara{Standard correlation clustering benchmarks:} To validate our method against traditional combinatorial algorithms in the standard transductive setting, we use unweighted graphs without node features. These include social, citation, and email networks from the SNAP and SuiteSparse collections.

\spara{Graph-level benchmarks:} To evaluate the inductive generalization of our model, we use datasets consisting in collections of multiple graphs. These include synthetic datasets (EXPWL1, a synthetic dataset designed to test Weisfeiler-Leman expressivity), molecular benchmarks (MUTAG, NCI1, OGBG-molhiv), protein association networks (OGBG-ppa), and social networks (GitHub-Stargazers, REDDIT-Binary).

\spara{Graph classification benchmarks:} For the graph pooling application, we employ a subset of the previous datasets, all of which have a corresponding graph classification task: EXPWL1, MUTAG, NCI1, and OGBG-molhiv.

\begin{table}
    \centering
    \caption{Statistics for all datasets. \textit{Standard Correlation Clustering} benchmark consist of individual graph instances, while the \textit{Graph-Level Tasks and Pooling} benchmark consist of collections of graphs (\# Nodes and \# Edges report average values). We report node features and number of labels exclusively of datasets used in pooling experiments. In the case of OGBG-ppa, we use only the first $30,000$ graphs of the dataset. 
    }
    \vspace{-4mm}
    \resizebox{\linewidth}{!}{
        \begin{tabular}{l@{}rrrrr}
        \toprule
        Dataset & \# Graphs & \# Nodes & \# Edges & \# Features & \# Labels \\
        \midrule
        \multicolumn{6}{l}{\textit{\textbf{Standard Correlation Clustering Benchmarks}}} \\
        polblogs\cite{davis2011university} & 1 & 1,224 & 16,715 & -- & -- \\
        ca-GrQc\cite{jure2014snap} & 1 & 5,241 & 14,484 & -- & -- \\
        ca-HepTh\cite{jure2014snap} & 1 & 9,875 & 25,973 & -- & -- \\
        ca-AstroPh\cite{jure2014snap} & 1 & 18,771 & 198,050 & -- & -- \\
        email-Enron\cite{jure2014snap} & 1 & 36,692 & 183,831 & -- & -- \\
        cond-mat-2005\cite{davis2011university} & 1 & 39,577 & 175,691 & -- & -- \\
        \midrule
        \multicolumn{6}{l}{\textit{\textbf{Graph-Level Tasks and Pooling Benchmarks}}} \\
        EXPWL1\cite{bianchi2023expressive} & 3,000 & 76.96 & 186.46 & 1 & 2 \\
        MUTAG\cite{morris2020tudataset} & 188 & 17.93 & 39.59 & 7 & 2 \\
        NCI1\cite{morris2020tudataset} & 4,110 & 29.87 & 64.60 & 37 & 2 \\
        REDDIT-BINARY\cite{morris2020tudataset} & 2,000 & 429.63 & 497.75 & -- & -- \\
        GitHub-Stargazers\cite{morris2020tudataset} & 12,725 & 113.79 & 234.64 & -- & -- \\
        OGBG-molhiv\cite{hu2020open} & 41,127 & 25.50 & 27.50 & 9 & 2 \\
        OGBG-ppa\cite{hu2020open} & 30,000 & 243.40 & 2,266.1 & -- & -- \\
        \bottomrule
        \end{tabular}
    }
    \label{tab:unified_datasets}
     \vspace{-5mm}
\end{table}

\subsection{Baselines}

We compare our framework against two distinct sets of state-of-the-art baselines, corresponding to the two primary problem settings. 

\spara{Correlation clustering baselines.} We compare with KwikCluster~\cite{ailon2008aggregating}, the seminal randomized 3-approximation algorithm; ModifiedPivot~\cite{behnezhad2025correlation}, which improves local cluster adjustments; and CFP (CoverFlipPivot)~\cite{bengali2023faster}, a combinatorial method for the generalized LambdaCC problem. For RQ1, we also compare with an exact Gurobi solver. 

\spara{Graph pooling baselines.} To assess the utility of Inductive CC as a pooling layer (\textsf{CCPool}), we compare against established pooling methods selected from the benchmark in \cite{bianchi2025torch}, including Deep Modularity Networks (DMON)~\cite{tsitsulin2023graph}, High-Order Spectral Clustering (HOSC)~\cite{duval2022higher}, Just-Balance Pooling (JBPool)~\cite{bianchi2023simplifying}, Asymmetric Cheeger Cut (ACC)~\cite{hansen2023total}, k-Maximal Independent Sets pooling (k-MIS)~\cite{bacciu2023generalizing}, and MaxCut Pooling~\cite{abatemaxcutpool}.

\subsection{Experimental Protocol}
\label{sec:protocol}

For all Correlation Clustering experiments, performance is measured using the solution cost as defined in Eq.~\ref{eq:node_wise_formulation}, normalized relative to the minimum cost achieved across all methods.

We evaluate our models, \textsf{NodeGNN} (transductive node-wise) and \textsf{LinkGNN} (inductive link-wise), across five distinct scenarios. For all experiments, we always report mean over different runs, together with the standard error of mean.

\spara{Approximation quality analysis:} To validate the correctness of our framework on tractable instances (i.e., small instances where exact optimization is feasible), we utilize synthetic graphs generated via Stochastic Block Models (SBMs)~\cite{HOLLAND1983109}. We fix the graph size at $|V|=36$ and divide nodes into $n$ equal-sized blocks. Connection probabilities are set to $0.8$ for intra-block edges and $0.1$ for inter-block edges. We vary $n$ from $2$ to $36$ (where $n=36$ approximates a random graph) and generate $10$ random instances per configuration, reporting the average performance across these $10$ instances. As the total graph size is fixed, increasing $n$ corresponds to a decrease in cluster size as well. We compare the cost of the GNN solution against the global optimum computed using the Gurobi exact solver.

\spara{Transductive optimization:}
We train and evaluate on the same graph instance to verify if the GNN can recover high-quality clustering solutions compared to combinatorial solvers. We run all methods $5$ times each to average against the stochasticity of the methods (i.e., the training algorithm of \textsf{NodeGNN}/\textsf{LinkGNN}, and the random pivot sampling which is used in different flavors in all the methods). All results are normalized with the minimum achieved in each dataset across all methods and iterations. We initialize nodes with random vectors, instead of structured-oriented embeddings (such as Laplacian positional encodings \cite{dwivedi2020generalization}), as these did not show to improve the models' results in preliminary experiments. Indeed, GNNs are typically more able to discern structural patterns when nodes are initialized with random features \cite{abboud2020surprising,sato2021random}.

\spara{Inductive generalization:}
To evaluate inductive clustering on unseen instances, we split the graph-level datasets into train (80\%), validation (10\%), and test (10\%) sets. For \textsf{LinkGNN}, we adapt Algorithm \ref{alg:link_wise_training} by using the validation set to select the optimal threshold for the test set. For each dataset, the models are trained on the graphs in the train set, and we report the performance on the graphs in the test set. As mentioned in \S \ref{sec:inductive_method}, since the optimal solution depends only on the graph structure, we discard original node features and use Node2Vec embeddings instead.
We also run the transductive methods Kwikcluster and ModifiedPivot on the test set to serve as lower bounds on solution quality. As in the transductive setup, we execute 5 runs per method using different random splits. Because these splits are identical across methods, we normalize each result by the minimum cost achieved on that specific split.

\spara{Graph pooling:}
We perform 10-fold cross-validation on the graph classification benchmarks. Here, \textsf{LinkGNN} is used as a pooling operator (``\textsf{CCPool}'') to coarsen the graph between GNN layers, selecting the clustering threshold to retain a fixed percentage of edges rather than minimizing clustering cost (\S \ref{sec:pooling}). Since the optimal pooling depends on both the downstream task and the graph structure, we use the downstream model embeddings as input for \textsf{CCPool}.
In line with standard benchmarks, we report average test accuracy, except for OGBG-molhiv, where ROC-AUC is used due to class imbalance.
For all the datasets and methods, we tune their corresponding hyperparameters to achieve a pooling ratio of at most 10\% of the original nodes, including the parameter $r$ of \textsf{CCPool}.

\spara{Ablation studies:} In \S \ref{sec:ablation_study}, we conduct comprehensive ablation and sensitivity studies to analyze our design choices. First, we vary model complexity: more complex configurations evaluate deeper architectures (2--3 GCN layers with nonlinearities between layers) or alternative convolutions (GIN \cite{xu2018how}, GraphSAGE \cite{hamilton2017inductive}, GAT \cite{velickovic2018graph}), while simpler variants remove message passing entirely by using a single linear layer ("Linear") or trainable node embeddings ("EMBOnly"). Second, we assess the operational impact of pivot sampling, embedding normalization, and final cluster selection. Finally, we perform a sensitivity analysis on both threshold selection and the embedding dimension $D$.

\subsection{Model Architectures and Hyperparameters}

We implement \textsf{NodeGNN} and \textsf{LinkGNN}, using a simplified architecture consisting of a single Graph Convolutional Network 
layer without nonlinearities. We do the same for \textsf{CCPool}. We provide additional implementation details in the following.

\spara{Architectures:} Both \textsf{NodeGNN} and \textsf{LinkGNN} are composed of a single GCN layer \cite{kipf2017semisupervised}, without nonlinearities. They adopt the approach described in \S \ref{sec:transductive} and \S \ref{sec:inductive_method}. We refer to the implementation of \textsf{LinkGNN} for pooling desribed in \S \ref{sec:pooling} as \textsf{CCPool}. Referring to the terminology of \cite{grattarola2022understanding}, \textsf{CCPool} technically uses \textsf{LinkGNN} as the "select" function of the pooling method, while for the "reduce" and "connect" operations we adopt standard choices: the aggregated feature matrix is given by a sum across the pooled nodes, and nodes are connected in the pooled graph if the two corresponding clusters were originally connected by at least an edge. In addition, we do not perform Pivot-based sampling in \textsf{CCPool}: since the graphs are already quite small, we do not require sampling to train efficiently. 

\spara{Hyperparameters:} In both \S \ref{sec:tCC_exp} and \S \ref{sec:iCC_exp}, we set the maximum number of clusters for \textsf{NodeGNN} to $K=\min \left(1e4,\, |V|\right)$, where $|V|$ is the number of nodes in the training graph. For \textsf{LinkGNN}, we use $512$ output channels in \S \ref{sec:tCC_exp} and $64$ in \S \ref{sec:iCC_exp}. Both models are trained for a maximum of $5000$ epochs, with early stopping triggered if the loss does not improve for $100$ consecutive epochs ($500$ epochs for \S \ref{sec:iCC_exp}). The configurations for features and sampling vary by section: \S \ref{sec:tCC_exp} uses $1000$ random pivots for sampling at each epoch alongside $512$ random features per node. Meanwhile, \S \ref{sec:iCC_exp} uses a batch size of $64$ for all datasets except ogbg-molhiv and ogbg-ppa, which use a batch size of $128$, and extracts one random pivot per each graph in the batch. For Node2Vec in \S \ref{sec:iCC_exp}, we use $128$ embedding dimensions and train for $100$ epochs with a walk length of $10$, a context size of $10$, and a learning rate of $0.01$. In \S \ref{sec:gurobi_Vs_gnn}, we adjust the setup to use $32$ random node features and $32$ output channels for \textsf{LinkGNN}. Finally, \S \ref{sec:pooling_exp} employs a GNN architecture consisting of two GIN convolutional layers \cite{xu2018how}, a pooling layer, a third GIN layer, a global pooling layer, and a three-layer MLP readout. Each layer contains $64$ hidden channels, except the final layer of the MLP, which has $32$. This model is trained for $1000$ epochs with a batch size of $64$, a learning rate of $1e-4$, and an early stopping patience of $100$ epochs.

\spara{Hardware and Compute Environment:} Experiments were conducted across two distinct computing environments based on workload requirements:  
\begin{squishlist}
    \item \textbf{Large-scale \& GPU:} Experiments on real-world datasets (\S \ref{sec:tCC_exp}, \ref{sec:iCC_exp}, \ref{sec:pooling_exp}) were executed on a machine with an NVIDIA A100 GPU (80GB VRAM) and an Intel Xeon Gold 6248R 15-core CPU.
    \item \textbf{Small-scale \& CPU:} Synthetic SBM experiments (\S \ref{sec:gurobi_Vs_gnn}) and all runs of the combinatorial baseline CFP were performed on an Apple M1 Pro processor (8-core CPU, 16GB RAM).
\end{squishlist}

\begin{figure}[t!]
    \centering
    \includegraphics[width=1.\linewidth]{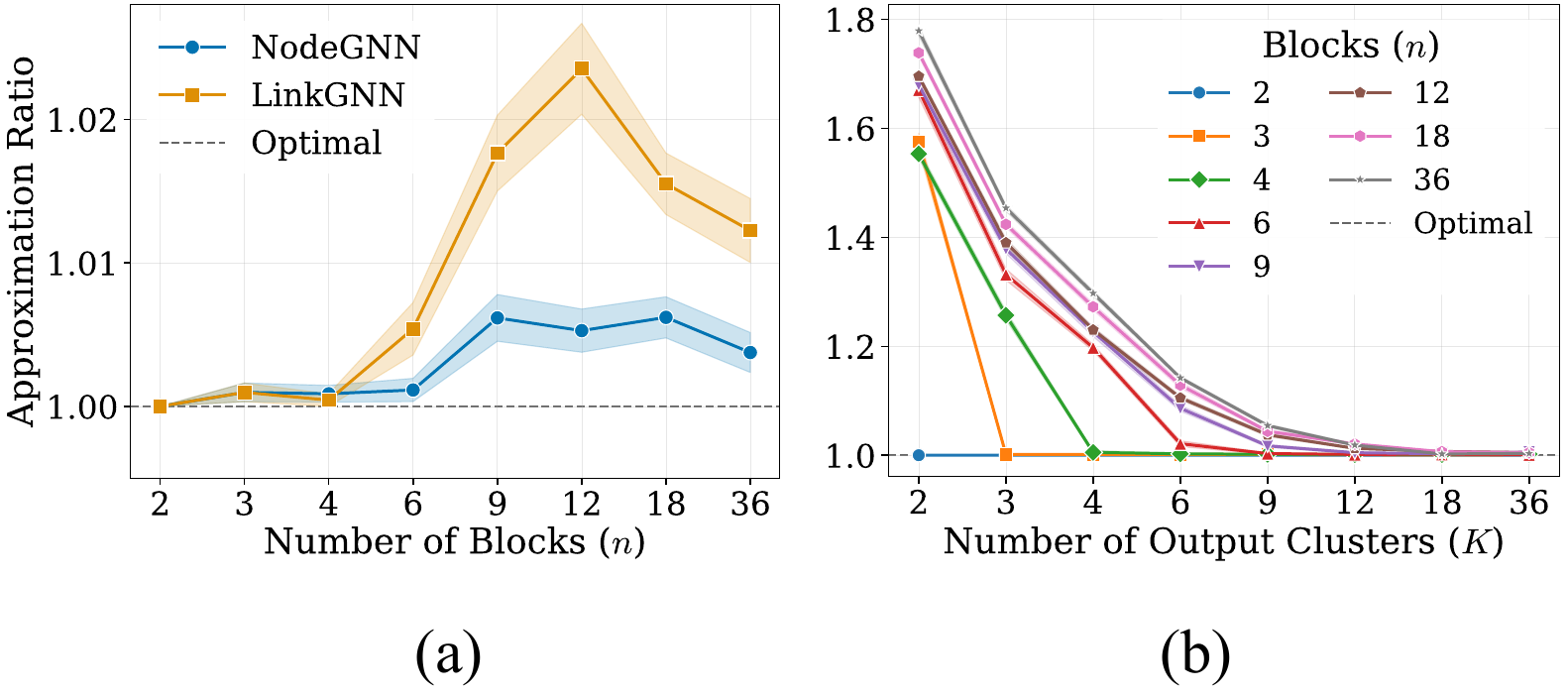}
    \vspace{-20pt}
    \caption{Approximation ratio analysis on synthetic Stochastic Block Model (SBM) graphs ($|V|=36$) relative to the optimal solution found by Gurobi. A ratio of 1.0 indicates an optimal solution. Shaded areas represent the standard error of the mean over 10 iterations. (a) Approximation ratio for \textsf{NodeGNN} ($K=36$) and \textsf{LinkGNN} as a function of the number of blocks $n$ (proxy for ground truth clusters). (b) Impact of the maximum cluster budget $K$ on the approximation quality of \textsf{NodeGNN} for varying block densities.}

    \label{fig:fig2}
    \vspace{-5mm}
\end{figure}

\section{Experimental Results and Discussion}
\label{sec:exp}
We report experimental results to evaluate the effectiveness, scalability, and versatility of the proposed framework. The study is organized around the four research questions in \S \ref{sec:exp_setup}. In the following subsections, we address each question in turn, using the datasets and protocols defined in Section~\ref{sec:exp_setup}.

\subsection{RQ1: Approximation Quality}
\label{sec:gurobi_Vs_gnn}

We compare GNN solutions against the exact Gurobi optimum on synthetic SBM graphs.
%
In Figure \ref{fig:fig2}a, we report the ratio between the GNN solution cost and the optimal value while increasing the number of blocks $n$ in the SBM.
The results show that both models achieve tight approximations, consistently staying within $1.05\times$ of the optimal cost.
However, the error for both methods generally increases with the number of blocks $n$. In particular, \textsf{LinkGNN} exhibits a larger error than \textsf{NodeGNN}, indicating that link-wise optimization introduces a greater approximation error.

While \textsf{LinkGNN} has no upper limit to the maximum number of output clusters, in \textsf{NodeGNN} this is controlled by the hyperparameter $K$. Figure \ref{fig:fig2}b reports the approximation ratio while varying $K$ in \textsf{NodeGNN}. 
It highlights a structural limitation of \textsf{NodeGNN}: the error spikes significantly if the cluster budget $K$ underestimates the true cluster count ($K < n$).
To ensure robustness when the optimal number of clusters
is unknown, \textsf{NodeGNN} requires setting $K \approx |V|$, which severely impacts efficiency.
This underscores \textsf{LinkGNN}'s advantage in resource-constrained scenarios involving large graphs with numerous clusters.

\addtolength{\tabcolsep}{-3.0pt}
\begin{table}[t!]
    \caption{
    Normalized cost comparison in the transductive setting.
    Values for each dataset are normalized relative to the minimum cost achieved across all methods (1.0 indicates the best performance).
    The horizontal line separates standard Correlation Clustering benchmarks (top) from node classification datasets (bottom). Best performance for each dataset is in bold, while second best performance is underlined. At the bottom, number of times a method appears as best or second-best out of all datasets.
    }
    \centering
    \vspace{-5pt}
    \resizebox{1.\columnwidth}{!}{
    \begin{tabular}{l|rrrr@{\;\;\;}r}
    \toprule
     & \multicolumn{1}{c}{NodeGNN} & \multicolumn{1}{c}{LinkGNN} & \multicolumn{1}{c}{KwikClus.} & \multicolumn{1}{c}{ModPivot} & \multicolumn{1}{c}{CFP} \\
    \midrule
    polblogs & \valpmb{1.00}{3e-4} & \valpmu{1.05}{1e-3} & \valpm{1.36}{0.04} & \valpm{1.12}{4e-4} & \valpm{1.27}{2e-3} \\
    GrQc & \valpmb{1.00}{1e-3} & \valpm{1.39}{4e-3} & \valpm{1.52}{0.02} & \valpmu{1.26}{0.01} & \valpm{2.04}{5e-3} \\
    HepTh & \valpmb{1.00}{1e-3} & \valpm{1.23}{2e-3} & {\valpm{1.48}{0.02}} & \valpmu{1.20}{4e-3} & \valpm{1.81}{2e-3} \\
    AstroPh & \valpmb{1.00}{7e-4} & \valpm{1.35}{7e-4} & \valpm{1.70}{0.12} & \valpmu{1.26}{0.01} & \valpm{1.76}{2e-3} \\
    Enron & \valpmu{1.14}{5e-3} & \valpmb{1.01}{2e-4} & \valpm{2.40}{0.82} & \valpmb{1.01}{4e-3} & \valpm{1.32}{8e-4} \\
    cond-mat & \valpmb{1.01}{5e-3} & \valpmu{1.09}{3e-4} & \valpm{1.25}{0.02} & \valpmb{1.01}{3e-3} & \valpm{1.45}{7e-4} \\
    \midrule
   \# 1st/2nd & 5/1 &  1/2 & 0/0 & 2/3 & 0/0 \\
    \bottomrule
    \end{tabular}
    }

    \label{tab:cc_benchmark}
    \vspace{-4mm}
\end{table}
\addtolength{\tabcolsep}{1pt}

\subsection{RQ2: Transductive Performance}
\label{sec:tCC_exp}

We first test our local optimization framework on a standard (non-inductive) setup. 
The results, reported in Table \ref{tab:cc_benchmark} shows that \textsf{NodeGNN} almost always obtains the solution with the minimum cost across competitors. By explicitly optimizing a relaxed Correlation Clustering objective for local minima, our method yields higher-quality solutions than competitors, which adopt combinatorial methods.
In contrast, \textsf{LinkGNN} often provides a worse approximation ratio than \textsf{NodeGNN} and Modified Pivot. However, on the large scale datasets email-Enron and cond-mat-2005, where the optimal number of clusters likely exceeds the $K=10,000$ cluster limit, \textsf{LinkGNN} ranks as best or second best method because its distance-based approach is not restricted by a maximum cluster count, allowing it to scale more effectively than \textsf{NodeGNN}.

\subsection{RQ3: Inductive Generalization and Efficiency}
\label{sec:iCC_exp}

\begin{figure}
    \centering
    \includegraphics[width=1.0\linewidth]{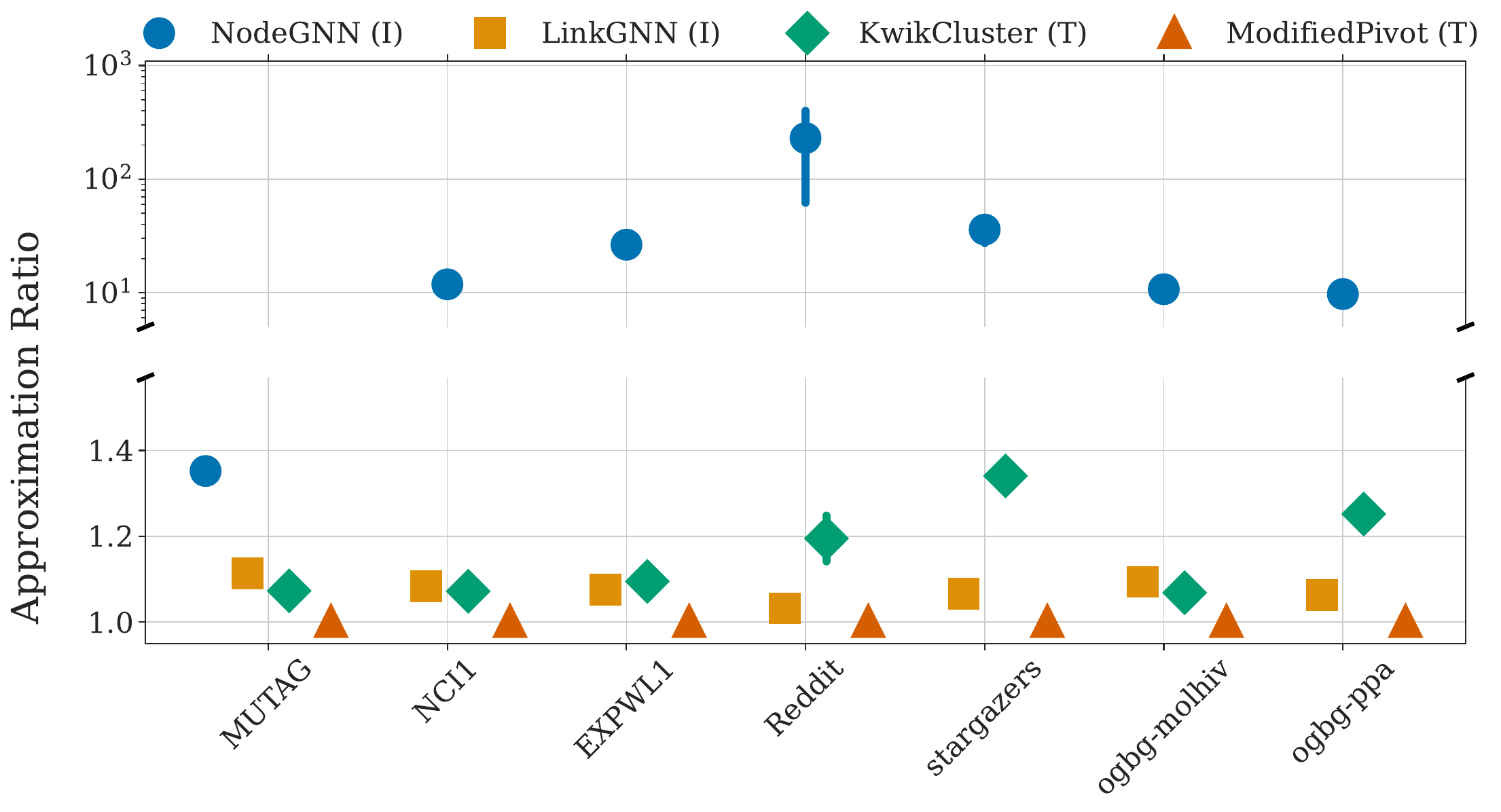}
    \vspace{-21pt}
    \caption{Comparison of methods in an inductive setting. Each point has error bars corresponding to the standard error of the mean, but are often too small. (I) was used to demark inductive methods, (T) for transductive ones. Values were normalized in each train/test split according to the minimum cost reached across methods.}
    \label{fig:inductive_feature_benchmark}
\vspace{-4mm}
\end{figure}

We now consider the inductive setup (\S \ref{sec:exp_setup}).
While previously \textsf{NodeGNN} usually performed better than \textsf{LinkGNN}, in this case (Figure~\ref{fig:inductive_feature_benchmark}) the cluster assignment learnt by \textsf{NodeGNN} is not well generalizable to a new graph. This performance collapse stems from architectural overfitting: by treating clusters as distinct classes, \textsf{NodeGNN} uses significantly more parameters than \textsf{LinkGNN}. Consequently, the model memorizes specific cluster identities rather than learning transferable structural motifs, causing it to fail on the novel distributions of unseen graphs.
\textsf{LinkGNN}, instead, obtains results better than KwikCluster on 4 out of 7 datasets, and in a few cases is almost on par than ModifiedPivot. By mapping nodes into a latent space where proximity reflects structural features and connectivity, \textsf{LinkGNN} captures universal clustering patterns instead of graph-specific assignments which allow to perform clustering inductively. 
Additionally, as shown in Table \ref{tab:inference_times}, since we are not re-training \textsf{LinkGNN} from scratch for each graph, at inference time is significantly faster than ModifiedPivot and KwikCluster.

\begin{table}[]
    \centering
    \caption{Inference time (in seconds) with the methods in the inductive setting of \S \ref{sec:iCC_exp}.}
    \vspace{-8pt}
    \resizebox{1.\columnwidth}{!}{
    \begin{tabular}{l|rrr}
    \toprule
    & LinkGNN & KwikCluster & ModifiedPivot \\
    \midrule

    MUTAG & \underline{\valpm{0.01}{2e-3}} & \textbf{\valpm{1e-3}{3e-5}} & \valpm{0.21}{0.7} \\
    NCI1 & \textbf{\valpm{0.04}{7e-3}} & \underline{\valpm{0.08}{7e-4}} & \valpm{5.70}{0.10} \\
    EXPWL1 & \textbf{\valpm{0.02}{7e-4}} & \underline{\valpm{0.36}{0.02}} & \valpm{11.73}{0.09} \\
    Reddit & \textbf{\valpm{0.02}{1e-3}} & \underline{\valpm{17.98}{1.85}} & \valpm{633.36}{106.97} \\
    stargazers & \textbf{\valpm{0.09}{9e-3}} & \underline{\valpm{6.37}{0.06}} & \valpm{314.00}{10.96} \\
    ogbg-molhiv & \textbf{\valpm{0.20}{0.05}} & \underline{\valpm{0.70}{4e-03}} & \valpm{46.72}{0.31} \\
    ogbg-ppa & \textbf{\valpm{0.52}{0.02}} & \underline{\valpm{181.54}{1.89}} & \valpm{2234.55}{11.17} \\
    \bottomrule
    \end{tabular}
    }

    \label{tab:inference_times}
\vspace{-4mm}
\end{table}

\subsection{RQ4: Application to Graph Pooling}
\label{sec:pooling_exp}

We adapt our inductive \textsf{LinkGNN} as a pooling technique (\S \ref{sec:pooling}). 

Table \ref{tab:pooling_results}, shows that \textsf{CCPool} delivers more consistent performance compared to state-of-the-art pooling mechanisms by combining structural flexibility with localized learning capabilities. On the EXPWL1 dataset, which evaluates the expressiveness of pooling architectures, dense poolers such as JBPool, HOSC, DMoN, and ACC are fundamentally constrained by a fixed maximum number of clusters $K$, often leading to failure in inductive clustering tasks. \textsf{CCPool}, like other sparse architectures such as MaxCut, SEP, and $k$-MIS, overcomes this limitation by dynamically adapting to new graphs through a variable number of output clusters.

Crucially, \textsf{CCPool} overcomes the traditional trade-offs that limit other sparse poolers. While sparse methods typically underperform on molecular datasets like MUTAG, NCI1, and OGBG-molhiv (or, like SEP, fail to complete training on larger datasets), where independent functional groups dominate molecular properties, \textsf{CCPool} consistently matches the performance of dense methods on these chemical tasks. This demonstrates that \textsf{CCPool} is uniquely capable of learning dataset-specific structures while simultaneously retaining the adaptive flexibility of a sparse pooling architecture, making it a more robust and versatile choice than its competitors.

\begin{table}[t]
    \caption{Accuracy of pooling methods on graph classification datasets. Accuracy is reported for all datasets except OGBG-molhiv, where ROC-AUC is used. For each dataset, best score is in bold and second-best is underlined. For OGBG-molhiv, SEP consistently exceeded our 48-hour time limit (OOT) without completing the training. Last column shows the average ranking of each method (lower is better).}
    \centering
    \vspace{-8pt}
    \resizebox{1.0\columnwidth}{!}{
    \begin{tabular}{l|rrrr|r}
    \toprule
    & EXPWL1 & MUTAG & NCI1 & molhiv & Avg. Rank \\
    \midrule
    CCPool & \textbf{\valpm{100.0}{0.0}} & {\valpm{84.2}{2.2}} & \underline{\valpm{78.0}{0.7}} & \textbf{\valpm{80.1}{0.8}} & \textbf{2.88} \\
    JBPool & \valpm{98.7}{0.3} & \valpm{82.0}{2.3} & \textbf{\valpm{78.9}{0.4}} & \valpm{77.6}{0.9} & 4.75 \\
    HOSC & \valpm{91.8}{0.7} & \underline{\valpm{85.1}{2.3}} & \valpm{77.8}{1.0} & \valpm{79.0}{0.6} & 5.25 \\
    DMoN & \valpm{96.3}{0.8} & \valpm{84.8}{1.9} & \valpm{77.5}{0.7} & \underline{\valpm{79.6}{0.8}} & 4.75 \\
    ACC & \valpm{93.3}{1.1} & \textbf{\valpm{91.6}{1.9}} & \underline{\valpm{78.0}{0.6}} &  \valpm{79.3}{0.8} & \underline{3.50} \\
    \textit{k}-MIS & \underline{\valpm{99.8}{0.1 }}& \valpm{81.6}{2.7} & \valpm{77.9}{0.8} &  {\valpm{79.5}{0.8}} & 4.50 \\
    MaxCut & \valpm{99.6}{0.2} & \valpm{76.1}{3.0} & \valpm{76.9}{0.4} & \valpm{78.0}{1.0} & 6.50\\
    SEP & \textbf{\valpm{100.0}{0.0}} & \valpm{85.0}{2.2} & \valpm{78.0}{0.5}& OOT & 3.88 \\

    \bottomrule
    \end{tabular}}
    \label{tab:pooling_results}
    \vspace{-5mm}
\end{table}

\subsection{Ablation Studies}
\label{sec:ablation_study}
As a further test on our framework, we analyze the performance of both \textsf{NodeGNN} and \textsf{LinkGNN} when altering the number and the type of layers on the Correlation Clustering benchmarks. We show the results in Table \ref{tab:num_layer_comparison} and Table \ref{tab:conv_type_comparison}, where the cost of the final solution is normalized with the minimum only across the methods used in this section (and not, e.g., the results of ModifiedPivot). We also analyze on REDDIT-binary additional ablations regarding our algorithm design choices (Table \ref{tab:ablation_study}), and performed a sensitivity analysis on the embedding dimension $D$ and the threshold selection (Table \ref{tab:thresh_and_dim}). In the sensitivity analysis, we are not using on the test set the threshold chosen during training, but perform a sweep to show how the solution quality varies according to the choice.  

\spara{Architecture Ablation:}
Our single-layer \textsf{NodeGNN} variant consistently performs best. As shown in Tables \ref{tab:num_layer_comparison} and \ref{tab:conv_type_comparison}, increasing layer depth, simplifying the architecture, or using alternative convolutions (except GraphSAGE) degrades performance. These results validate the efficacy of GNNs for this framework; constraining the model to a single layer keeps the parameter count manageable, which is crucial since the output dimension is fixed to $K=\min(1e4,|V|)$.  For \textsf{LinkGNN}, results are more nuanced: a single convolutional layer is optimal for all datasets except ca-AstroPh, where the embedding-only variant ("EMBOnly") excels. However, EMBOnly is transductive and cannot generalize to unseen graphs. GCN and GAT yield comparable performance, so we choose GCN for architectural consistency with \textsf{NodeGNN}. In this link-wise setup, architectural choices are less impactful because the parameter count is significantly lower, though the optimization still benefits from structural graph information and simpler models.

\begin{table}[t]
    \centering
    \caption{Ablation study on model complexity of \textsf{NodeGNN} and \textsf{LinkGNN}, comparing the clustering cost (Eq. \ref{eq:node_wise-loss}). The best model on each row is in bold; costs are normalized for each dataset across all methods used in the table.}
    \vspace{-3mm}
    \resizebox{\linewidth}{!}{
    \begin{tabular}{c|ccccc}
        \toprule
        \
        Dataset &  EMBOnly & Linear & 1 Layer & 2 Layers & 3 Layers \\
        \midrule
        \multicolumn{6}{c}{\textbf{NodeGNN}} \\
        polblogs & \valpm{1.06}{0.03} & \valpm{1.11}{5e-4} & \valpmb{1.00}{2e-4} & \valpm{3.75}{2.38} & \valpm{25.89}{4.63}\\
        GrQc & \valpm{2.36}{0.24} & \valpm{2.35}{2e-3} & \valpmb{1.00}{6e-4} & \valpm{90.00}{9.79} & \valpm{2224.25}{0.00} \\
        HepTh & \valpm{1.74}{0.02} & \valpm{1.64}{1e-3} & \valpmb{1.00}{9e-4} & \valpm{3.66}{0.18} & \valpm{2965.73}{0.00} \\
        AstroPh & \valpm{1.65}{6e-3} & \valpm{1.64}{2e-4} & \valpmb{1.00}{1e-3} & \valpm{2.16}{0.04} & \valpm{1371.64}{0.00} \\
        Enron & \valpm{1.50}{2e-3} & \valpm{1.56}{2e-3} & \valpmb{1.12}{5e-3} & \valpm{5.71}{2.80} & \valpm{4003.65}{0.00} \\
        cond-mat & \valpm{1.80}{0.01} & \valpm{1.88}{1e-3} & \valpmb{1.01}{5e-3} & \valpm{4.06}{0.31} & \valpm{5643.22}{0.00} \\
        \midrule
        \addlinespace
        \multicolumn{6}{c}{\textbf{LinkGNN}} \\
        polblogs & \valpm{1.05}{1e-5} & \valpm{1.06}{8e-4} & \valpmb{1.04}{7e-4} & \valpm{1.07}{3e-3} & \valpm{1.13}{0.01} \\
        GrQc & \valpm{1.33}{0.03} & \valpm{1.98}{3e-3} & \valpmb{1.32}{1e-3} & \valpm{1.43}{0.01} & \valpm{1.46}{0.02} \\
        HepTh & \valpm{1.40}{0.02} & \valpm{1.58}{0.00} & \valpmb{1.23}{3e-4} & \valpm{1.29}{1e-3} & \valpm{1.30}{6e-3} \\
        AstroPh & \valpmb{1.29}{4e-3} & \valpm{1.54}{0.00} & \valpm{1.38}{2e-4} & \valpm{1.40}{5e-3} & \valpm{1.37}{9e-3} \\
        Enron & \valpm{1.09}{1e-4} & \valpm{1.09}{1e-6} & \valpmb{1.00}{2e-4} & \valpm{1.04}{4e-4} & \valpm{1.14}{0.03} \\
        cond-mat & \valpm{1.23}{1e-3} & \valpm{1.27}{0.00} & \valpmb{1.05}{2e-3} & \valpm{1.11}{3e-3} & \valpm{1.12}{0.01} \\
            \bottomrule
    \end{tabular}
    }
    \label{tab:num_layer_comparison}
\vspace{-4mm}
\end{table}

\begin{table}[t]
    \centering
    \caption{Ablation study on the message passing operation by comparing clustering costs. In all models, we use a single message passing layer. As in Table \ref{tab:num_layer_comparison}, the costs are normalized for each dataset across all models used in the table. Best performance for each line is in bold.}
    \vspace{-4mm}
    \resizebox{1.0\columnwidth}{!}{
    \begin{tabular}{c|cccc}
        \toprule
        Dataset & GCN & GraphSAGE & GIN & GAT \\
        \midrule
        \multicolumn{5}{c}{\textbf{NodeGNN}} \\
        polblogs & \valpmb{1.00}{2e-4} & \valpm{1.05}{1e-3} & \valpm{1.39}{0.05} & \valpm{1.68}{0.10} \\
        GrQc & \valpmb{1.00}{6e-4} & \valpm{1.30}{4e-3} & \valpm{30.00}{2.99} & \valpm{2.18}{0.08} \\
        HepTh & \valpmb{1.00}{9e-4} & \valpm{1.25}{1e-3} & \valpm{55.47}{8.22} & \valpm{2.18}{0.05} \\
        AstroPh & \valpmb{1.00}{1e-3} & \valpm{1.08}{9e-4} & \valpm{42.88}{7.06} & \valpm{2.26}{0.10} \\
        Enron & \valpmb{1.12}{5e-3} & \valpm{7.07}{0.87} & \valpm{1,056.9}{351.4} & \valpm{6.26}{0.58} \\
        cond-mat & \valpmb{1.01}{5e-3} & \valpm{1.33}{5e-3} & \valpm{3,350.3}{583.1} & \valpm{3.01}{0.09} \\
        \midrule
        \addlinespace
        \multicolumn{5}{c}{\textbf{LinkGNN}} \\
        polblogs & \valpmb{1.04}{7e-4} & \valpm{1.06}{1e-3} & \valpm{1.06}{2e-3} & \valpm{1.08}{8e-3} \\
        GrQc & \valpm{1.32}{1e-3} & \valpm{1.37}{4e-3} & \valpm{1.50}{0.03} & \valpmb{1.31}{9e-3} \\
        HepTh & \valpmb{1.23}{3e-4} & \valpm{1.28}{7e-3} & \valpm{1.36}{0.01} & \valpm{1.24}{3e-3} \\
        AstroPh & \valpm{1.38}{2e-4} & \valpm{1.36}{2e-4} & \valpm{1.37}{2e-3} & \valpmb{1.34}{9e-3} \\
        email-Enron & \valpmb{1.00}{2e-4} & \valpm{5.99}{0.50} & \valpm{1.12}{0.02} & \valpm{1.08}{0.01} \\
        cond-mat & \valpm{1.05}{2e-3} & \valpm{1.11}{5e-3} & \valpm{1.19}{4e-3} & \valpmb{1.04}{1e-3} \\
    \end{tabular}}
    \label{tab:conv_type_comparison}
    \vspace{-4mm}

\end{table}

\spara{Inductive Algorithm Ablation:} We observe  from Table \ref{tab:ablation_study} that each design choice affect the quality of our solution:
\begin{squishlist}
    \item Pivot Sampling: It drastically reduces memory and training time without degrading quality (as mentioned in \S \ref{sec:pivot}).
    \item L2 Normalization: Normalization of the embeddings allows the method to find proper solutions for our CC problem relaxation by constraining distances in a [0,1] range, and speeds up convergence 5x. Additionally, it has theoretical grounding in contrastive learning~\cite{wang2020understanding}. We hypothesize that also in this case it grants convergence on alignment (similar intra-cluster nodes are grouped together) and uniformity (evenly distributed dissimilar clusters). While a low embedding dimension $D$ on large graphs could cause small distinct clusters to overlap.
    \item Rounding clusters: Substituting our clustering inference heuristic (Algorithm \ref{alg:inductive_inference}) with the Charikar rounding from \cite{charikar2005clustering} increases clustering costs, as this method assume a different CC relaxation. 
\end{squishlist}

\begin{table}[]
\centering
\caption{Ablation study on inductive algorithm design choices. We compute the cost ratio relative to the baseline configuration obtained without any ablation ("Baseline")).}
\vspace{-3mm}
\label{tab:ablation_study}
\begin{tabular}{l|ccc}
\toprule
Ablation & Approx. Ratio & Train time (s) & GPU (MB) \\
\midrule
Baseline & \valpm{1.0}{2e-03} & \valpm{190}{9} & \valpm{17751}{1091} \\
No pivot & \valpm{1.0}{7e-04} & \valpm{8166}{523} & \valpm{75709}{1287} \\
No L2 norm. & \valpm{1.0}{2e-03} & \valpm{918}{18} & \valpm{17751}{1091} \\
Charikar clusters & \valpm{1.1}{3e-03} & \valpm{197}{9} & \valpm{17748}{1092} \\
\bottomrule
\end{tabular}
\vspace{-3mm}
\end{table}

\spara{Sensitivity Analysis:} From Table \ref{tab:thresh_and_dim} we see that the choice of embedding dimension $D$ does not have a strong impact on the quality of the solution as long as it is large enough. Instead, while the exact threshold impacts quality, the training-optimized value (typically $t \approx 0.90$) transfers robustly to unseen graphs. 

\begin{table}[]
    \centering
    \caption{Approximation ratio depending on embedding dimension $D$ and threshold $t$ selection. The ratio is computed with respect to the minimum across all configurations.}
    \vspace{-3mm}
    \begin{tabular}{l|ccccc}
    \toprule
   $D$ & $t=0.25$ & $t=0.50$ & $t=0.90$ & $t=0.95$ & $t=1.00$ \\
    \midrule
    $16$ & \valpm{232.50}{22.74} & \valpm{67.18}{7.62} & \valpm{1.06}{0.02} & \valpm{1.03}{8e-03} & \valpm{1.09}{9e-03} \\
    $64$ & \valpm{222.51}{23.73} & \valpm{61.47}{7.63} & \valpm{1.00}{2e-03} & \valpm{1.03}{9e-03} & \valpm{1.09}{9e-03} \\
    $256$ & \valpm{221.53}{23.98} & \valpm{61.75}{7.86} & \valpm{1.01}{5e-03} & \valpm{1.03}{9e-03} & \valpm{1.09}{9e-03} \\
    \bottomrule
    \end{tabular}

    \label{tab:thresh_and_dim}

\vspace{-5mm}
\end{table}

\section{Conclusions}

This work introduces Inductive Correlation Clustering, a novel extension of Correlation Clustering that learns a parametrized function to handle unseen graph instances. We solve this problem by establishing \textsf{CC-GNN}, a local optimization framework for Correlation Clustering which takes advantage of the inductive properties of GNNs, alongside a pivot-based sampling strategy to ensure scalable and efficient training on large graphs. In particular, we consider two alternative formulations, which we dub \textsf{NodeGNN} and \textsf{LinkGNN}. Through comprehensive experiments, we demonstrate the efficacy of \textsf{NodeGNN} in achieving superior solution quality in transductive settings with respect to competitors; \textsf{LinkGNN}, instead, successfully generalizes to new graphs providing solutions of quality comparable to transductive baselines, while spending significantly less inference time. Finally, we demonstrate an application of Inductive Correlation Clustering to graph pooling. We adapt \textsf{LinkGNN} into a learnable pooling mechanism, \textsf{CCPool}, and showcase its competitiveness with state-of-the-art techniques in graph classification tasks. In summary, this research combines classic combinatorial optimization and deep learning techniques, offering robust and scalable solutions for clustering graph data in large and dynamic batches.  Future work will explore inductive applications of Correlation Clustering extensions (e.g. overlapping clusters or diversity constraints), and identify improved rounding algorithms.

\begin{acks}
Part of this work was done while F.P.N. was visiting Bowling Green State University. 
F.P.N. and A.K. gratefully acknowledge the Ohio Supercomputer Center (OSC) for providing the computing resources and facilities used in conducting part of the experiments.
\end{acks}

\section*{GenAI Usage Disclosure}

Generative AI tools were used in a assistive capacity in the preparation of this work: as a coding assistant during the implementation of the experimental pipeline, and for proofreading and editorial polishing of the manuscript. The research ideas, the problem formulation, the design of the proposed methods, the experimental protocol, and the analysis and interpretation of the results are entirely the authors' own. All AI-assisted code and text were reviewed and validated by the authors, who take full responsibility for the content of this paper.

\FloatBarrier


\bibliographystyle{ACM-Reference-Format}
\bibliography{sample-base}



\end{document}